\documentclass[10pt,twocolumn,letterpaper]{article}

\usepackage{iccv}
\usepackage{times}
\usepackage{epsfig}
\usepackage{graphicx}
\usepackage{amsmath}
\usepackage{amssymb}
\usepackage[normalem]{ulem}

\usepackage{cite}
\usepackage{multirow}

\usepackage[pagebackref=true,breaklinks=true,letterpaper=true,colorlinks,bookmarks=false]{hyperref}

\iccvfinalcopy 

\begin{document}
\newcommand{\twofigures}[3]{
            \centerline{{\includegraphics[width=#3]{#1}}~~{\includegraphics[width=#3]{#2}}}}

\newcommand{\threefiguresh}[4]{
            \centerline{{\includegraphics[height=#4]{#1}}{\includegraphics[height=#4]{#2}}{\includegraphics[height=#4]{#3}}}}

\newcommand{\threefigures}[4]{
            \centerline{{\includegraphics[width=#4]{#1}}~~{\includegraphics[width=#4]{#2}}~~{\includegraphics[width=#4]{#3}}}}

\newcommand{\threefigureslbl}[5]{
            \centerline{(#5){\includegraphics[width=#4]{#1}}~~{\includegraphics[width=#4]{#2}}~~{\includegraphics[width=#4]{#3}}}}

\newcommand{\lthreefigures}[4]{
            \centerline{{\includegraphics[width=#4]{#1}}{\includegraphics[width=#4]{#2}}{\includegraphics[width=#4]{#3}}
			\makebox[0.33\columnwidth][c]{(a)}\makebox[0.33\columnwidth][c]{(b)}\makebox[0.33\columnwidth][c]{(c)}}}

\newcommand{\fourfigures}[5]{
            \centerline{{\includegraphics[width=#5]{#1}}~~{\includegraphics[width=#5]{#2}}~~{\includegraphics[width=#5]{#3}}~~{\includegraphics[width=#5]{#4}}}}

\newcommand{\fourfigurescaption}[9]{
            \centerline{{\includegraphics[width=#5]{#1}}~~{\includegraphics[width=#5]{#2}}~~{\includegraphics[width=#5]{#3}}~~{\includegraphics[width=#5]{#4}}}
            \makebox[#5][c]{#6}\makebox[#5][c]{#7}\makebox[#5][c]{#8}\makebox[#5][c]{#9}}
            
\newcommand{\fivefigures}[6]{
            \centerline{{\includegraphics[width=#6]{#1}}~~{\includegraphics[width=#6]{#2}}~~{\includegraphics[width=#6]{#3}}~~{\includegraphics[width=#6]{#4}}~~{\includegraphics[width=#6]{#5}}}}

\newcommand{\fivefigurescaption}[6]{
            \centerline{{\includegraphics[width=#6]{#1}}~~{\includegraphics[width=#6]{#2}}~~{\includegraphics[width=#6]{#3}}~~{\includegraphics[width=#6]{#4}}~~{\includegraphics[width=#6]{#5}}}\makebox[#6][c]{Ground truth~~~~~~~~}\makebox[#6][c]{~~CUT~\cite{keuper2015motion}}\makebox[#6][c]{~~FST~\cite{papazoglou2013fast}}\makebox[#6][c]{~~~~MP-Net-Frame~\cite{tokmakov2016learning}}\makebox[#6][c]{~~~~~~~~~Ours}}   

\newcommand{\sixfigures}[7]{
            \centerline{{\includegraphics[width=#7]{#1}}~~{\includegraphics[width=#7]{#2}}~~{\includegraphics[width=#7]{#3}}~~{\includegraphics[width=#7]{#4}}~~{\includegraphics[width=#7]{#5}}~~{\includegraphics[width=#7]{#6}}}}
            
\newcommand{\sixfiguresdots}[7]{
            \centerline{{\includegraphics[width=#7]{#1}}~~{\includegraphics[width=#7]{#2}}~~{\includegraphics[width=#7]{#3}}~~{...}~~{\includegraphics[width=#7]{#4}}~~{\includegraphics[width=#7]{#5}}~~{\includegraphics[width=#7]{#6}}}}

\newcommand{\threefigurescaption}[7]{
            \centerline{{\includegraphics[width=#4]{#1}}~~{\includegraphics[width=#4]{#2}}~~{\includegraphics[width=#4]{#3}}}
		     \makebox[#4][c]{#5}\makebox[#4][c]{#6}\makebox[#4][c]{#7}}

\newcommand{\comment}[1]{}

\newcommand{\twofigurescaption}[6]{
            \centerline{{\includegraphics[width=#3]{#1}}~~{\includegraphics[width=#3]{#2}}}
            \makebox[#6][c]{#4}\makebox[#6][c]{#5}}

\newcommand{\todo}[1]{{\bf TODO:} #1}

\graphicspath{{figs/}}

\title{Learning Video Object Segmentation with Visual Memory}

\author{Pavel Tokmakov
\and
Karteek Alahari\vspace{0.3cm}\\
\large{Inria\thanks{Univ.\ Grenoble Alpes, Inria, CNRS, Grenoble INP, LJK, 38000 Grenoble, France.}\vspace{-0.2cm}}
\and
Cordelia Schmid\\
}

\maketitle

\begin{abstract}
This paper addresses the task of segmenting moving objects in unconstrained
videos. We introduce a novel two-stream neural network with an explicit memory
module to achieve this. The two streams of the network encode spatial and
temporal features in a video sequence respectively, while the memory module
captures the evolution of objects over time. The module to build a ``visual
memory'' in video, i.e., a joint representation of all the video frames, is
realized with a convolutional recurrent unit learned from a small number of
training video sequences. Given a video frame as input, our approach assigns
each pixel an object or background label based on the learned spatio-temporal
features as well as the ``visual memory'' specific to the video, acquired
automatically without any manually-annotated frames. The visual memory is
implemented with convolutional gated recurrent units, which allows to propagate
spatial information over time. We evaluate our method extensively on two
benchmarks, DAVIS and Freiburg-Berkeley motion segmentation datasets, and show
state-of-the-art results. For example, our approach outperforms the top method
on the DAVIS dataset by nearly 6\%. We also provide an extensive ablative
analysis to investigate the influence of each component in the proposed
framework.
\end{abstract}

\section{Introduction}
\label{sec:intro}
\begin{figure}[t]
\begin{center}
\twofigures{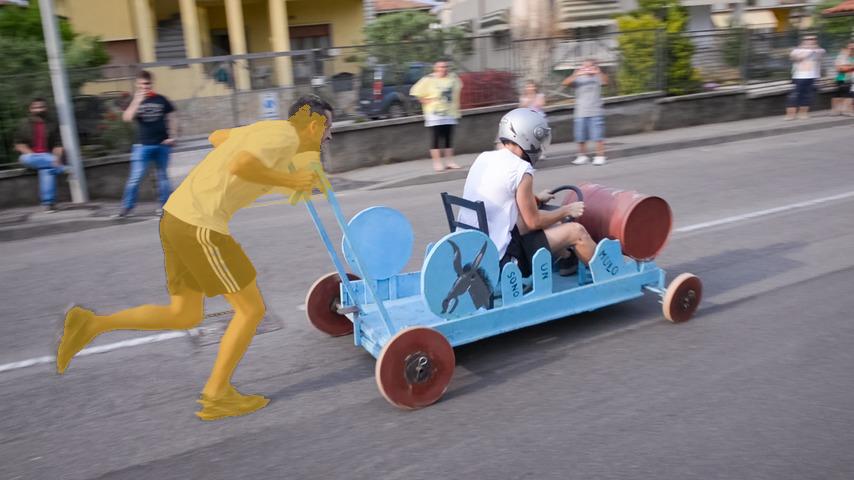}{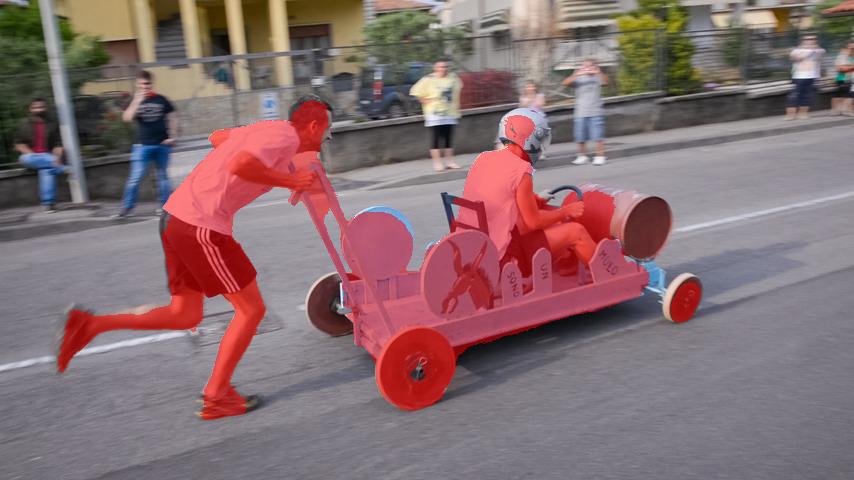}{0.5\linewidth}\vspace{0.1cm}
\twofigures{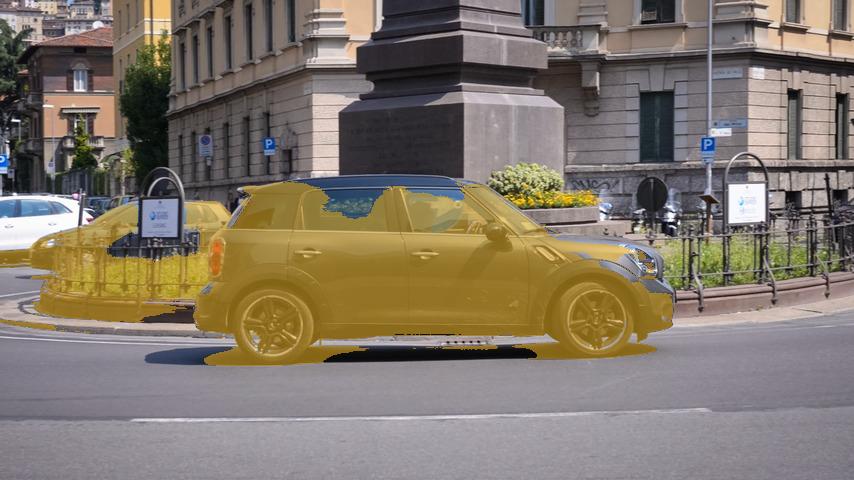}{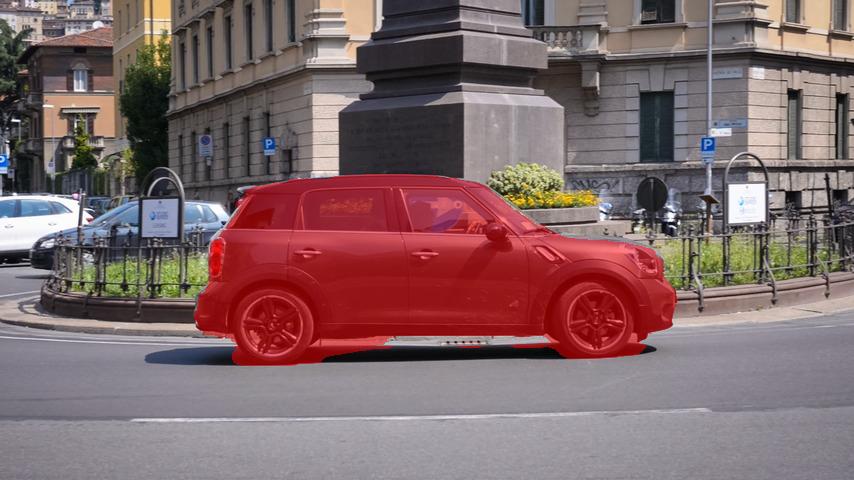}{0.5\linewidth}
\end{center}
\vspace{-0.7cm}\caption{Sample results on the DAVIS dataset. Segmentations
produced by MP-Net~\cite{tokmakov2016learning} (left) and our approach (right),
overlaid on the video frame.\vspace{-0.5cm}}
\label{fig:teaser}
\end{figure}

Video object segmentation is the task of extracting spatio-temporal regions
that correspond to object(s) moving in at least one frame in the video
sequence. The top-performing methods for this
problem~\cite{papazoglou2013fast,Faktor14} continue to rely on hand-crafted
features and do not leverage a learned video representation, despite the
impressive results achieved by convolutional neural networks (CNN) for other
vision tasks, e.g., image segmentation~\cite{pinheiro2016learning}, object
detection~\cite{ren2015faster}. Very recently, there have been attempts to
build CNNs for video object
segmentation~\cite{tokmakov2016learning,Caelles17,Khoreva16}. They are indeed
the first to use deep learning methods for video segmentation, but suffer from
various drawbacks. For example,~\cite{Caelles17,Khoreva16} rely on a
manually-segmented subset of frames (typically the first frame of the video
sequence) to guide the segmentation pipeline. Our previous
work~\cite{tokmakov2016learning} relies solely on optical flow between pairs of
frames to segment independently moving objects in a video, making it
susceptible to errors in flow estimation. It also can not extract objects if
they stop moving. Furthermore, none of these methods has a mechanism to {\it
memorize} relevant features of objects in a scene. In this paper, we propose a
novel framework to address these issues; see sample results in
Figure~\ref{fig:teaser}.

We present a two-stream network with an explicit memory module for video object
segmentation (see Figure~\ref{fig:model}). The memory module is a convolutional
gated recurrent unit (GRU) that encodes the spatio-temporal evolution of
object(s) in the input video sequence. This spatio-temporal representation used
in the memory module is extracted from two streams---the appearance stream
which describes static features of objects in the video, and the temporal
stream which captures motion cues.

The appearance stream is the DeepLab network~\cite{chen2014semantic} pretrained
on the PASCAL VOC segmentation dataset and operates on individual video frames.
The temporal one is a motion prediction network~\cite{tokmakov2016learning}
pretrained on the synthetic FlyingThings3D dataset and takes optical flow
computed from pairs of frames as input, as shown in Figure~\ref{fig:model}. The
two streams provide complementary cues for object segmentation. With these
spatio-temporal CNN features in hand, we train the convolutional GRU component
of the framework to learn a {\it visual memory} representation of object(s) in
the scene. Given a frame $t$ from the video sequence as input, the network
extracts its spatio-temporal features and then: (i)~computes the segmentation
using the memory representation aggregated from all frames previously seen in
the video, and (ii)~updates the memory unit with features from $t$. The
segmentation is improved further by processing the video bidirectionally in the
memory unit, with our {\it bidirectional convolutional GRU}.

The contributions of the paper are two-fold. First, we present an approach for
moving object segmentation in unconstrained videos that does not require any
manually-annotated frames in the input video (see \S\ref{sec:overview}).  Our
network architecture incorporates a memory unit to capture the evolution of
object(s) in the scene (see \S\ref{sec:recurrent}). To our knowledge, this is
the first recurrent network based approach to accomplish the video segmentation
task. It helps address challenging scenarios where the motion patterns of the
object change over time; for example, when an object in motion stops to move,
abruptly, and then moves again, with potentially a different motion pattern.
Second, we present state-of-the-art results on two video object segmentation
benchmarks, namely DAVIS~\cite{Perazzi16} and Freiburg-Berkeley motion
segmentation (FBMS) dataset~\cite{ochs2014segmentation} (see \S\ref{sec:soa}).
Additionally, we provide an extensive experimental analysis, with ablation
studies to investigate the influence of all the components of our framework
(see \S\ref{sec:abl}) and visualize the internal states of our memory unit (see
\S\ref{sec:gru}). We will make the source code and the models available online.  

\section{Related work}
\label{sec:rel}

\noindent \textbf{Video object segmentation.}
Several approaches have been proposed over the years to accomplish the task of
segmenting objects in video. One of the more successful ones presented
in~\cite{brox2010object} clusters pixels spatio-temporally based on motion
features computed along individual point trajectories. Improvements to this
framework include dense trajectory-level segmentation~\cite{ochs2011object}, an
alternative clustering method~\cite{keuper2015motion}, and detection of
discontinuities in the trajectory spectral
embedding~\cite{fragkiadaki2012video}. These trajectory based approaches lack
robustness in cases where feature matching fails.

An alternative to using trajectories is formulating the segmentation problem as
a foreground-background classification
task~\cite{papazoglou2013fast,lee2011key,wang2015saliency}. These methods first
estimate a region~\cite{papazoglou2013fast,wang2015saliency} or
regions~\cite{lee2011key}, which correspond(s) to the foreground object, and
then use them to compute foreground and background appearance models. The final
object segmentation is obtained by integrating these appearance models with
other cues, e.g., saliency maps~\cite{wang2015saliency}, shape
estimates~\cite{lee2011key}, pairwise constraints~\cite{papazoglou2013fast}.
Variants to this framework have introduced occlusion relations to compute a
layered video segmentation~\cite{taylor2015causal}, and long-range interactions
to group re-occurring regions in video~\cite{Faktor14}. Two methods from this
class of segmentation approaches~\cite{papazoglou2013fast,Faktor14} show a good
performance on the DAVIS benchmark. While our proposed method is similar in
spirit to this class of approaches, in terms of formulating segmentation as a
classification problem, we differ from previous work significantly. We propose
an integrated approach to learn appearance and motion features and update them
with a memory module, in contrast to estimating an initial region heuristically
and then propagating it over time. Our robust model outperforms all these
methods~\cite{papazoglou2013fast,wang2015saliency,lee2011key,taylor2015causal,Faktor14},
as shown in Section~\ref{sec:soa}.

\begin{figure*}[th]
\begin{center}
\includegraphics[width=2.10\columnwidth]{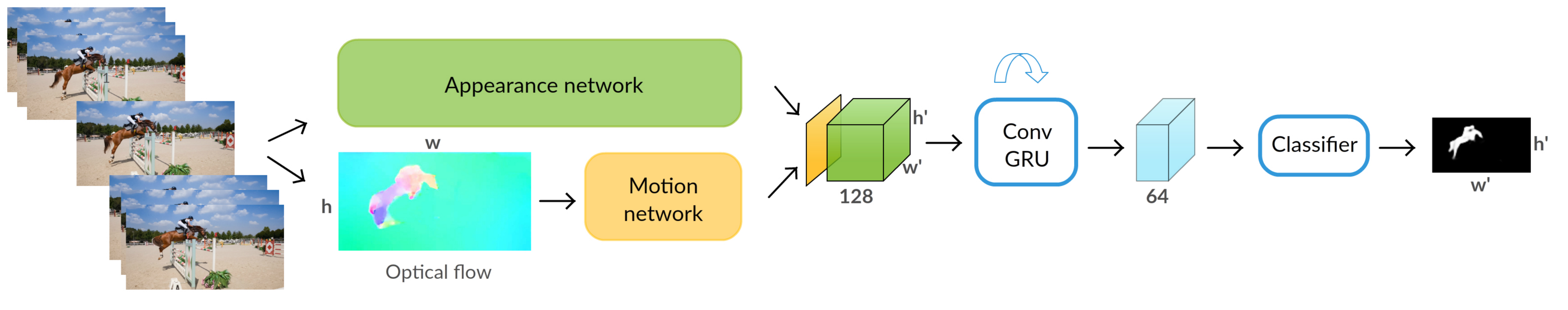}
\end{center}
\vspace{-0.3cm}
\caption{Overview of our segmentation approach. Each video frame is processed
by the appearance (green) and the motion (yellow) networks to produce an
intermediate two-stream representation. The ConvGRU module combines this with
the learned visual memory to compute the final segmentation result. The width
(w') and height (h') of the feature map and the output are $\text{w}/8$ and
$\text{h}/8$ respectively.}
\vspace{-0.3cm}
\label{fig:model}
\end{figure*}

Video object segmentation is also related to the task of segmenting objects in
motion, irrespective of camera motion. Two recent methods to address this task
use optical flow computed between pairs of
frames~\cite{Bideau16,tokmakov2016learning}. Classical methods in perspective
geometry and RANSAC-based feature matching are used in~\cite{Bideau16} to
estimate moving objects from optical flow. It achieved state-of-the-art
performance on a subset of the Berkeley motion segmentation (BMS)
dataset~\cite{brox2010object}, but lacks robustness due to a heuristic
initialization, as shown in the evaluation on DAVIS in
Table~\ref{tbl:soadavis}. Our previous approach
MP-Net~\cite{tokmakov2016learning} learns to recognize motion patterns in a
flow field. This frame-based approach is the state of the art on DAVIS and is
on par with~\cite{Bideau16} on the BMS subset.  Despite its excellent
performance, MP-Net is limited by its frame-based nature and also overlooks
appearance features of objects. These issues are partially addressed in a
heuristic post-processing step with objectness cues
in~\cite{tokmakov2016learning}. Nevertheless, the approach fails to extract
objects if they stop moving, i.e., if no motion cues are present. We use MP-Net
as the temporal stream of our  approach (see Figure~\ref{fig:model}). We show a
principled way to integrate this stream with appearance information and a new
visual memory module based on convolutional gated recurrent units (ConvGRU). As
shown in Table~\ref{tbl:mpn}, our approach outperforms MP-Net. 

Very recently, two CNN-based methods for video object segmentation were
proposed~\cite{Khoreva16,Caelles17}. Starting with CNNs pre-trained for image
segmentation, they find objects in video by fine-tuning on the first frame in
the sequence. Note that this setup, referred to as semi-supervised
segmentation, is very different from the more challenging unsupervised case we
address in this paper, where no manually-annotated frames are available for the
test video. Furthermore, these two CNN architectures are primarily developed
for images, and do not model temporal information in video. We, on the other
hand, propose a recurrent network specifically for the video segmentation task.

\noindent \textbf{Recurrent neural networks (RNNs).}
RNN~\cite{hopfield1982neural,rumelhart86} is a popular model for tasks defined
on sequential data. Its main component is an internal state that allows to
accumulate information over time. The internal state in classical RNNs is
updated with a weighted combination of the input and the previous state, where
the weights are learned from training data for the task at hand. Long
short-term memory (LSTM)~\cite{hochreiter1997long} and gated recurrent unit
(GRU)~\cite{Cho14} architectures are improved variants of RNN, which partially
mitigate the issue of vanishing
gradients~\cite{pascanu2013difficulty,Hochreiter98}. They introduce gates with
learnable parameters, to update the internal state selectively, and can
propagate gradients further through time. 

Recurrent models, originally used for text and speech recognition,
e.g.,~\cite{graves2013speech,mikolov2010recurrent}, are becoming increasingly
popular for visual data. Initial work on vision tasks, such as image
captioning~\cite{donahue2015long}, future frame
prediction~\cite{srivastava2015unsupervised} and action
recognition~\cite{NgHVVMT15}, has represented the internal state of the
recurrent models as a 1D vector---without encoding any spatial information.
LSTM and GRU architectures have been extended to address this issue with the
introduction of
ConvLSTM~\cite{xingjian2015convolutional,patraucean2015spatio,finn2016unsupervised}
and ConvGRU~\cite{ballas2015delving} respectively. In these convolutional
recurrent models the state and the gates are 3D tensors and the weight vectors
are replaced by 2D convolutions. These models have only recently been applied
to vision tasks, such as video frame
prediction~\cite{finn2016unsupervised,patraucean2015spatio,xingjian2015convolutional},
action recognition and video captioning~\cite{ballas2015delving}.

In this paper, we employ a visual memory module based on a convolutional GRU
(ConvGRU) and show that it is an effective way to encode the spatio-temporal
evolution of objects in video for segmentation. Further, to fully benefit from
all the frames in a video sequence, we apply the recurrent model
bidirectionally~\cite{graves2005framewise,graves2013hybrid}, i.e., apply two
identical model instances on the sequence in forward and backward directions,
and combine the predictions for each frame. 

\section{Approach}
\label{sec:overview}
Our model takes video frames together with their estimated optical flow as
input, and outputs binary segmentations of moving objects, as shown in
Figure~\ref{fig:model}. We target the most general form of this task, wherein
objects are to be segmented in the entire video if they move in at least one
frame. The proposed model is comprised of three key components: appearance and
motion networks, and a visual memory module.

\noindent \textbf{Appearance network.}
The purpose of the appearance stream is to produce a high-level encoding of a
frame that will later aid the visual memory module in forming a representation
of the moving object. It takes an RGB frame as input and produces a $128 \times
\text{w}/8 \times \text{h}/8$ feature representation (shown in green in
Figure~\ref{fig:model}). This encodes the semantic content of the scene. We use
a state-of-the-art CNN for this stream, namely the largeFOV version of the
DeepLab network~\cite{chen2014semantic}. This network relies on dilated
convolutions~\cite{chen2014semantic}, which preserve a relatively high spatial
resolution of features, and also incorporate context information in each
pixel's representation. It is pretrained on a semantic segmentation dataset,
PASCAL VOC 2012~\cite{pascalvoc2012}, resulting in features that can
distinguish objects from background as well as from each other---a crucial
aspect for the video object segmentation task. We extract features from the fc6
layer of the network, which has a feature dimension of 1024 for each pixel.
This feature map is further passed through two $1 \times 1$ convolutional
layers, interleaved with \textit{tanh} nonlinearities, to reduce the dimension
to 128. These layers are trained together with ConvGRU (see
\S\ref{sec:implement} for details).

\noindent \textbf{Motion network.}
For the temporal stream we employ MP-Net~\cite{tokmakov2016learning}, a CNN
pretrained for the motion segmentation task. It is trained to estimate
independently moving objects (i.e., irrespective of camera motion) based on
optical flow computed from a pair of frames as input (shown in yellow in
Figure~\ref{fig:model}). This stream produces a $\text{w}/4 \times \text{h}/4$
motion prediction output, where each value represents the likelihood of the
corresponding pixel being in motion. Its output is further downsampled by a
factor 2 (in w and h) to match the dimensions of the appearance stream output.

The intuition behind using two streams is to benefit from their complementarity
for building a strong representation of objects that evolves over time. For
example, both appearance and motion networks are equally effective when an
object is moving in the scene, but as soon as it becomes stationary, the motion
network can not estimate the object, unlike the appearance network. We leverage
this complementary nature, as done by two-stream networks for other vision
tasks~\cite{simonyan2014two}. Note that our approach is not specific to the
particular networks described above, but is in fact a general framework for
video object segmentation. As shown is the Section~\ref{sec:abl}, its
components can easily be replaced with other networks, providing scope for
future improvement.

\noindent \textbf{Memory module.}
The third component, i.e., a visual memory module based on convolutional gated
units (ConvGRU), takes the concatenation of appearance and motion stream
outputs as its input. It refines the initial estimates from these two networks,
and also memorizes the appearance and location of objects in motion to segment
them in frames where: (i)~they are static, or (ii)~motion prediction fails; see
the example in Figure~\ref{fig:teaser}. The output of this ConvGRU memory
module is a $64 \times \text{w}/8 \times \text{h}/8$ feature map obtained by
combining the two-stream input with the internal state of the memory module, as
described in detail in Section~\ref{sec:recurrent}. We further improve the
model by processing the video bidirectionally; see Section~\ref{sec:bidirec}.
The output from the ConvGRU module is processed by a $1 \times 1$ convolutional
layer and softmax nonlinearity to produce the final pixelwise segmentation
result. These layers are trained together with ConvGRU, as detailed in
Section~\ref{sec:implement}.

\section{Visual memory module}
\label{sec:recurrent}
\begin{figure}[t]
\begin{center}
\includegraphics[width=\columnwidth]{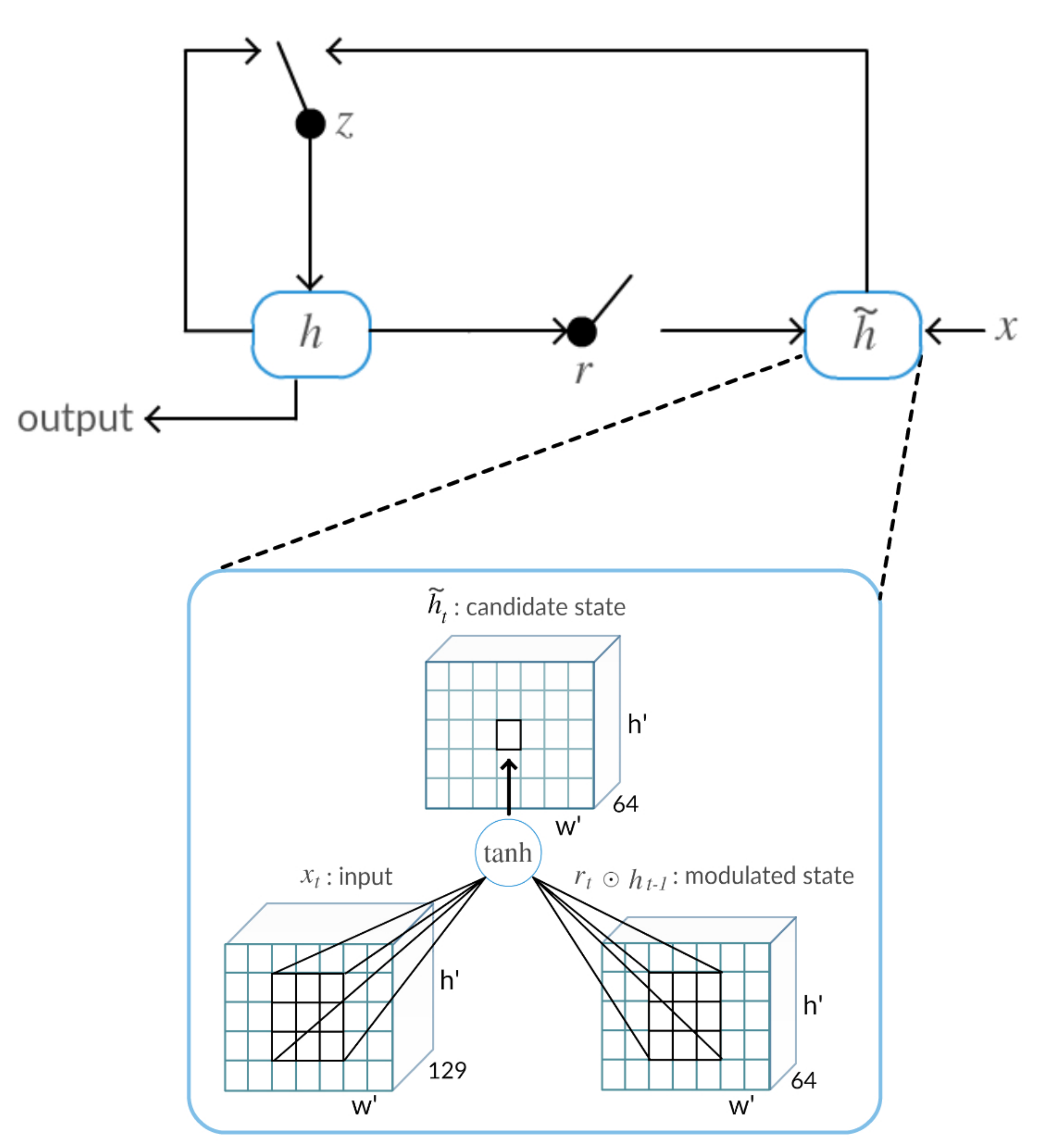}
\end{center}
\vspace{-0.3cm}
\caption{Illustration of ConvGRU with details for the candidate hidden state
module, where  $\tilde{h}_t$ is computed with two convolutional operations and
a $\tanh$ nonlinearity.}
\vspace{-0.3cm}
\label{fig:lstm}
\end{figure}

The key component of the ConvGRU module is the state matrix $h$, which encodes
the visual memory. For frame $t$ in the video sequence, ConvGRU uses the
two-stream representation $x_t$ and the previous state $h_{t-1}$ to compute the
new state $h_t$. The dynamics of this computation are guided by an update gate
$z_t$, a forget gate $r_t$. The states and the gates are 3D tensors, and can
characterize spatio-temporal patterns in the video, effectively memorizing
which objects move, and where they move to. These components are computed with
convolutional operators and nonlinearities as follows.
\begin{eqnarray}
	z_t &=& \sigma(x_t * w_{xz} + h_{t-1} * w_{hz} + b_{z}), \label{eqn:update} \\
	r_t &=& \sigma(x_t * w_{xr} + h_{t-1} * w_{hr} + b_{r}), \label{eqn:reset} \\
	\tilde{h}_t &=& \tanh(x_t * w_{x\tilde{h}} + r_t \odot h_{t-1} * w_{h\tilde{h}} + b_{\tilde{h}}), \label{eqn:candmem} \\
	h_t &=& (1 - z_t) \odot h_{t-1}  + z_t \odot \tilde{h}_t, \label{eqn:state}
\end{eqnarray}
where $\odot$ denotes element-wise multiplication, $*$ represents a
convolutional operation, $\sigma$ is the sigmoid function, $w$'s are learned
transformations, and $b$'s are bias terms.

The new state $h_t$ in (\ref{eqn:state}) is a weighted combination of the
previous state $h_{t-1}$ and the candidate memory $\tilde{h}_t$. The update
gate $z_t$ determines how much of this memory is incorporated into the new
state. If $z_t$ is close to zero, the memory represented by $\tilde{h}_t$ is
ignored. The reset gate $r_t$ controls the influence of the previous state
$h_{t-1}$ on the candidate memory $\tilde{h}_t$ in (\ref{eqn:candmem}), i.e.,
how much of the previous state is let through into the candidate memory. If
$r_t$ is close to zero, the unit forgets its previously computed state
$h_{t-1}$.

The gates and the candidate memory are computed with convolutional operations
over $x_t$ and $h_{t-1}$ shown in equations
(\ref{eqn:update}-\ref{eqn:candmem}). We illustrate the computation of the
candidate memory state $\tilde{h}_t$ in Figure~\ref{fig:lstm}.  The state at
$t-1$, $h_{t-1}$, is first multiplied (element-wise) with the reset gate $r_t$.
This modulated state representation and the input $x_t$ are then convolved with
learned transformations, $w_{h\tilde{h}}$ and $w_{x\tilde{h}}$ respectively,
summed together with a bias term $b_{\tilde{h}}$, and passed through a $\tanh$
nonlinearity. In other words, the visual memory representation of a pixel is
determined not only by the input and the previous state at that pixel, but also
its local neighborhood. Increasing the size of the convolutional kernels allows
the model to handle spatio-temporal patterns with larger motion.

The update and reset gates, $z_t$ and $r_t$, are computed in an analogous
fashion using a sigmoid function instead of $\tanh$. Our ConvGRU applies a
total of six convolutional operations at each time step. All the operations
detailed here are fully differentiable, and thus the parameters of the
convolutions ($w$'s and $b$'s) can be trained in an end-to-end fashion with
back propagation through time~\cite{werbos1990backpropagation}. In summary, the
model learns to combine appearance features of the current frame with the
memorized video representation to refine motion predictions, or even fully
restore them from the previous observations in case a moving object becomes
stationary.

\subsection{Bidirectional processing}
\label{sec:bidirec}
\begin{figure}[t]
\begin{center}
\includegraphics[width=1.0\columnwidth]{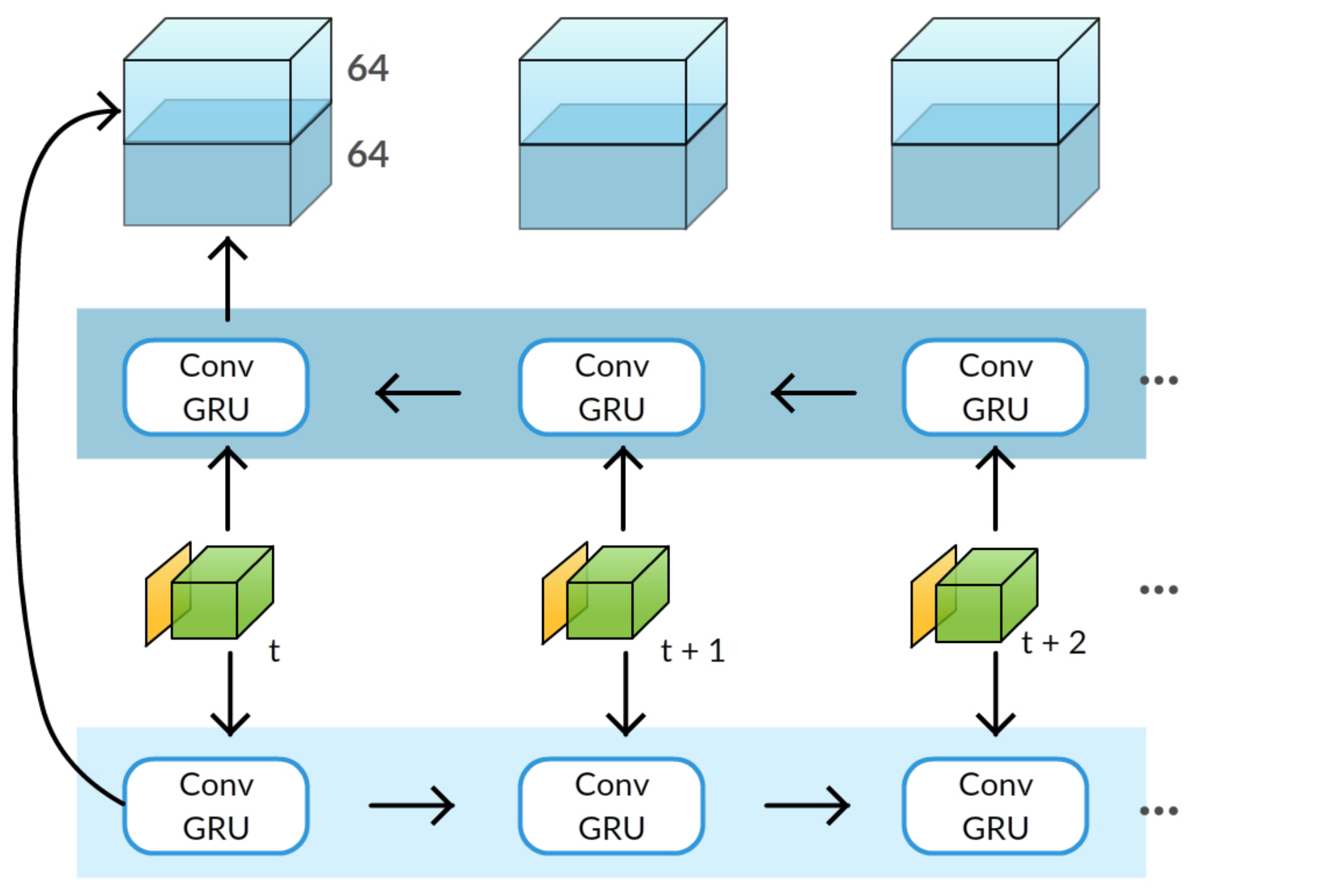}
\end{center}
\vspace{-0.3cm}
\caption{Illustration of the bidirectional processing with our ConvGRU module.}
\vspace{-0.3cm}
\label{fig:bidir}
\end{figure}

Consider an example where an object is stationary at the beginning of a video
sequence, and starts to move in the latter frames. Our approach described so
far, which processes video frames sequentially (in the forward direction), can
not segment the object in the initial frames.  This is due to the lack of prior
memory representation of the object in the first frame. We improve our
framework with a bidirectional processing step, inspired by the application of
recurrent models bidirectionally in the speech
domain~\cite{graves2005framewise,graves2013hybrid}.

The bidirectional variant of our ConvGRU is illustrated in
Figure~\ref{fig:bidir}. It is composed of two ConvGRU instances with identical
learned weights, which are run in parallel. The first one processes frames in
the forward direction, starting with the first frame (shown at the bottom in
the figure). The second instance process frames in the backward direction,
starting with the last video frame (shown at the top in the figure). The
activations from these two directions are concatenated at each time step, as
shown in the figure, to produce a $128 \times w/8 \times h/8$ output. It is
then passed through a $3 \times 3$ convolutional layer to finally produce a $64
\times w/8 \times h/8$ for each frame. Pixel-wise segmentation is then obtained
with a final $1 \times 1$ convolutional layer and a softmax nonlinearity, as in
the unidirectional case.

Bidirectional ConvGRU is used both in training and in testing, allowing the
model to learn to aggregate information over the entire video. In addition to
handling cases where objects move in the latter frames, it improves the ability
of the model to correct motion prediction errors. As discussed in the
experimental evaluation, bidirectional ConvGRU improves segmentation
performance by nearly 3\% on the DAVIS dataset (see Table~\ref{tbl:abl}). The
influence of bidirectional processing is more prominent on the FBMS dataset, where objects can be static in the beginning of a video,
with 5\% improvement over the unidirectional variant.

\subsection{Training}
\label{sec:train}
We train our visual memory module with the back propagation through time
algorithm~\cite{werbos1990backpropagation}, which unrolls the recurrent network
for $n$ time steps and keeps all the intermediate activations to compute the
gradients. Thus, our ConvGRU model, which has 6 internal convolutional layers,
trained on a video sequence of length $n$, is equivalent to a $6n$ layer CNN
for the unidirectional variant, or $12n$ for the bidirectional model at
training time. This memory requirement makes it infeasible to train the
whole model, including appearance and motion streams, end-to-end. We resort to
using pretrained versions of the appearance and motion networks and train the
ConvGRU.

We use the training split of the DAVIS dataset~\cite{Perazzi16} for learning
the ConvGRU weights. Objects move in all the frames in this dataset, which
biases the memory module towards the presence of an uninterrupted motion
stream. This results in the ConvGRU learned from this data failing, when an
object stops to move in a test sequence. We augment the training data to
simulate such {\it stop-and-go} scenarios and thus learn a more robust model
for realistic videos.

We create additional training sequences, where ground truth moving object
segmentation (instead of responses from the motion network) is provided for all
the frames, except for the last five frames, which are duplicated, simulating a case where objects stop moving. No motion input is used for these
last five frames. These artificial examples are used in place of the regular ones for a fixed fraction of iterations. Replacing motion stream predictions with ground truth segmentations for these sequences allows to decouple the task of motion mistake correction from the task of object tracking, which simplifies the learning. Given that ground truth segmentation determines the loss for training,
i.e., it is used for all the frames, ConvGRU explicitly memorizes the moving
object in the initial part of the sequence, and then segments it in frames
where motion is missing. We do a similar training set augmentation by duplicating the first five frames in a batch, to simulates the cases where an object is static in the beginning of a video.

\section{Experiments}

\subsection{Datasets and evaluation}
We use four datasets in the experimental analysis: DAVIS for training and test,
FBMS and SegTrack-v2 only for test, and FT3D for training a variant of our
approach.

\begin{table}[t]
\begin{center}
\begin{tabular}{l|l|c}
\hline
Aspect & Variant & Mean IoU  \\
\hline
  \multicolumn{2}{l|}{Ours (fc6, ConvGRU, Bidir, DAVIS)}  & 70.1    \\
  \hline
  \multirow{4}{*}{App stream} & no  & 43.5  \\
   & RGB  & 58.3  \\
   & 2-layer CNN  & 60.9 \\
   & DeepLab fc7  & 69.8  \\
   & DeepLab conv5  & 67.7  \\
  \hline
  App pretrain & ImageNet only  & 64.1  \\
  \hline
  Motion stream & no  & 59.6  \\
  \hline
  \multirow{3}{*}{Memory module} & ConvRNN & 68.7  \\
  & ConvLSTM  & 68.9  \\  
  & no  &  64.1 \\  
    \hline
  Bidir processing & no & 67.2  \\
  \hline
     \multirow{2}{*}{Train data} & FT3D GT Flow & 55.3  \\
   & FT3D LDOF Flow & 59.6   \\
  \hline
\end{tabular}
\vspace{0.1cm}
\caption{Ablation study on the DAVIS validation set showing variants of
appearance and motion streams and memory module. ``Ours'' refers to the model
using fc6 appearance features together with a motion stream, and a 
bidirectional ConvGRU trained on DAVIS.}
\label{tbl:abl}
\vspace{-0.6cm}
\end{center}
\end{table}

\vspace{0.3cm}\noindent \textbf{DAVIS.}
It contains 50 full HD videos with accurate pixel-level annotation in all the
frames~\cite{Perazzi16}. The annotations correspond to the task of video object
segmentation. Following the 30/20 training/validation split provided with the
dataset, we train on the 30 sequences, and test on the 20 validation videos. We
also follow the standard protocol for evaluation from~\cite{Perazzi16}, and
report intersection over union, F-measure for contour accuracy and temporal
stability.

\vspace{0.3cm}\noindent\textbf{FBMS.}
The Freiburg-Berkeley motion segmentation dataset~\cite{ochs2014segmentation}
is composed of 59 videos with ground truth annotations in a subset of the
frames. In contrast to DAVIS, it has multiple moving objects in several videos
with instance-level annotations. Also, objects may move only in a fraction of
the frames, but they are annotated in frames where they do not exhibit
independent motion. The dataset is split into training and test sets. 
Following the standard protocol on this dataset~\cite{keuper2015motion}, we do
not train on any of these sequences, and evaluate separately for both
with precision, recall and F-measure scores. We also convert
instance-level annotation to binary ones by merging all the foreground
labels into a single category, as in~\cite{taylor2015causal}.

\vspace{0.3cm}\noindent\textbf{SegTrack-v2.}
It contains 14 videos with instance-level moving object annotations in all the
frames. We convert these annotations into a binary form for evaluation and use
intersection over union as a performance measure.

\vspace{0.3cm}\noindent\textbf{FT3D.}
The FlyingThings3D dataset~\cite{Mayer16} consists of 2250 synthetic videos for
training, composed of 10 frames, where objects are in motion along random
trajectories in rendered scenes. Ground truth optical flow, depth, camera
parameters, and instance segmentations are provided by~\cite{Mayer16}, and the
ground truth motion segmentation is available from~\cite{tokmakov2016web}.

\subsection{Implementation details}
\label{sec:implement}
We train our model by minimizing binary cross-entropy loss using
back-propagation through time and RMSProp~\cite{rmsprop} with a learning rate
of $10^{-4}$. The learning rate is gradually decreased after every epoch. The
weight decay is set to $0.005$. Initialization of all the convolutional layers,
except for those inside the ConvGRU, is done with the standard \textit{xavier}
method~\cite{glorot2010understanding}. We clip the gradients to the $[-50, 50]$
range before each parameter update, to avoid numerical
issues~\cite{graves2013generating}. We form batches of size 14 by randomly
selecting a video, and a subset of 14 consecutive frames in it. Random cropping
and flipping of the sequences is also performed for data augmentation. Our full
model uses $7 \times 7$ convolutions in all the ConvGRU operations. The weights
of the two $1 \times 1$ convolutional (dimensionality reduction) layers in the
appearance network and the final $1 \times 1$ convolutional layer following the
memory module are learned jointly with the memory module. The model is trained
for 30000 iterations and the proportion of batches with additional sequences
(see Section~\ref{sec:train}) is set to 20\%.

Our final model uses a fully-connected CRF~\cite{krahenbuhl2011efficient} to
refine boundaries in a post-processing step. The parameters of this CRF are
taken from~\cite{tokmakov2016learning}. In the experiments where objectness is
used, it is also computed according to~\cite{tokmakov2016learning}. We use
LDOF~\cite{Brox11a} for optical flow estimation and convert the raw flow to
flow angle field, as in~\cite{tokmakov2016learning}. We used the code and
the trained models for MP-Net available at~\cite{tokmakov2016web}. Our
method is implemented in the Torch framework and will be made available online.
Many sequences in FBMS are several hundred frames long and do not fit into GPU
memory during evaluation. We apply our method in a sliding window fashion in
such cases, with a window of 130 frames and a step size of 50.

\subsection{Ablation study}
\label{sec:abl}
\begin{table}[t]
\begin{center}
\begin{tabular}{l|c}
\hline
Method & Mean IoU  \\
\hline
  Ours         & 70.1 \\
  Ours + CRF   & 75.9 \\
  \hline
  MP-Net       & 53.6 \\
  MP-Net + Obj & 63.3 \\  
  MP-Net + Obj + FST (MP-Net-V) & 55.0 \\
  MP-Net + Obj + CRF (MP-Net-F) & 70.0 \\
\hline
\end{tabular}
\vspace{0.1cm}
\caption{Comparison to MP-Net~\cite{tokmakov2016learning} variants on the DAVIS
validation set. ``Obj'' refers to the objectness cues used
in~\cite{tokmakov2016learning}. MP-Net-V(ideo) and MP-Net-F(rame) are the variants
of MP-Net which use FST~\cite{papazoglou2013fast} and CRF respectively, in
addition to objectness.}
\label{tbl:mpn}
\vspace{-0.6cm}
\end{center}
\end{table}

\begin{table*}[t]
\begin{center}
\begin{small}
\begin{tabular}{l | c | rrrrrrrcr}
\hline
\multicolumn{2}{c|}{Measure} & PCM~\cite{Bideau16} & CVOS~\cite{taylor2015causal} & KEY~\cite{lee2011key} & MSG~\cite{brox2010object} & NLC~\cite{Faktor14} & CUT~\cite{keuper2015motion} & FST~\cite{papazoglou2013fast} & MP-Net-F~\cite{tokmakov2016learning} & Ours \\
\hline
\multirow{3}{*}{$\mathcal{J}$} & Mean  & 40.1 & 48.2 & 49.8 & 53.3 & 55.1 & 55.2 & 55.8 & 70.0 & 75.9  \\
& Recall & 34.3 & 54.0 & 59.1 & 61.6  & 55.8 & 57.5 & 64.9 & 85.0 & 89.1  \\
& Decay & 15.2 & 10.5 & 14.1 & 2.4 & 12.6 & 2.3 & 0.0 & ~~1.4 & 0.0  \\
\hline
\multirow{3}{*}{$\mathcal{F}$} & Mean  & 39.6 & 44.7 & 42.7 & 50.8 & 52.3 & 55.2 & 51.1 & 65.9 & 72.1  \\
& Recall  & 15.4 & 52.6 & 37.5 & 60.0 & 51.9 & 61.0 & 51.6 & 79.2 & 83.4  \\
& Decay  & 12.7 & 11.7 & 10.6 & 5.1 & 11.4 & 3.4 & 2.9 & ~~2.5 & 1.3  \\
\hline
$\mathcal{T}$ & Mean  & 51.3 & 24.4 & 25.2 & 29.1 & 41.4 & 26.3 & 34.3 & 56.3 & 25.5  \\
\hline
\end{tabular}
\end{small}
\vspace{0.1cm}
\caption{Comparison to state-of-the-art methods on DAVIS with intersection over
union ($\mathcal{J}$), F-measure ($\mathcal{F}$), and temporal stability
($\mathcal{T}$).}
\label{tbl:soadavis}
\vspace{-0.4cm}
\end{center}
\end{table*}

\begin{figure*}[t]
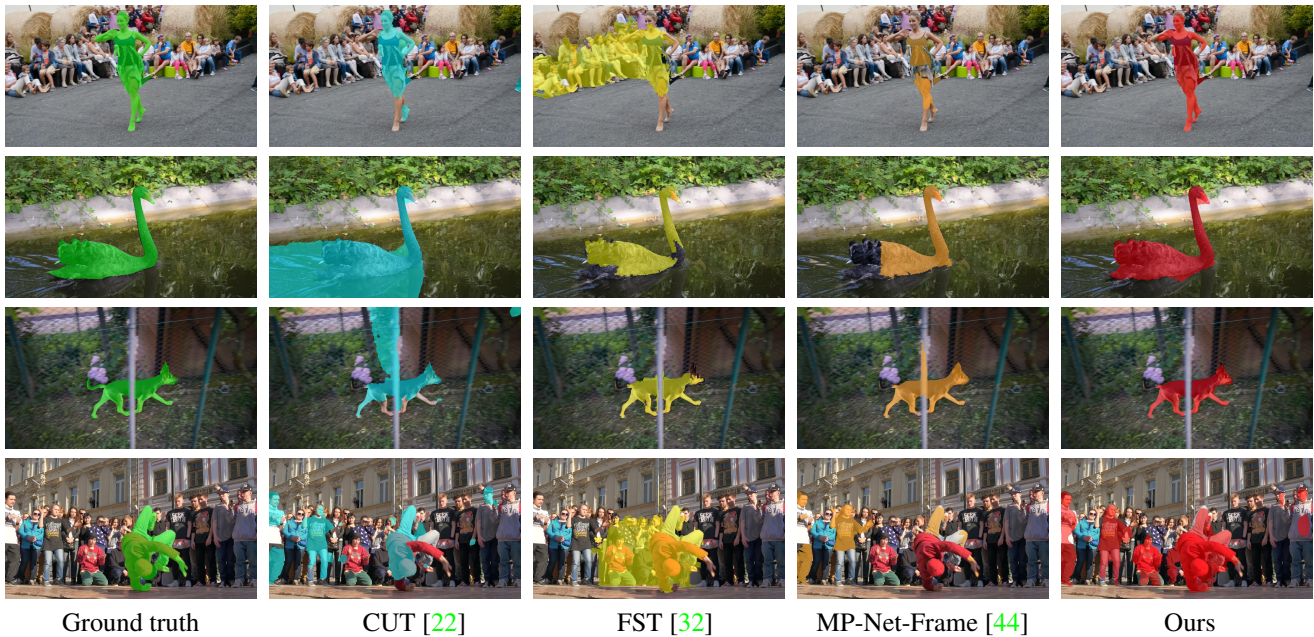

\begin{center}
\fivefigures{soa/dance-twirl/gt}{soa/dance-twirl/nlc}{soa/dance-twirl/vito}{soa/dance-twirl/mpnet}{soa/dance-twirl/ours}{0.40\columnwidth}\vspace{0.1cm}
\fivefigures{soa/blackswan/gt}{soa/blackswan/nlc}{soa/blackswan/vito}{soa/blackswan/mpnet}{soa/blackswan/ours}{0.40\columnwidth}\vspace{0.1cm}
\fivefigures{soa/libby/gt}{soa/libby/nlc}{soa/libby/vito}{soa/libby/mpnet}{soa/libby/ours}{0.40\columnwidth}\vspace{0.1cm}
\fivefigurescaption{soa/breakdance/gt}{soa/breakdance/nlc}{soa/breakdance/vito}{soa/breakdance/mpnet}{soa/breakdance/ours}{0.40\columnwidth}
\end{center}
\vspace{-0.3cm}\caption{Qualitative comparison with top-performing methods on
DAVIS. Left to right: ground truth, results of CUT~\cite{keuper2015motion},
FST~\cite{papazoglou2013fast}, MP-Net-Frame~\cite{tokmakov2016learning}, and our method.}
\vspace{-0.4cm}
\label{fig:davis}
\end{figure*}

Table~\ref{tbl:abl} demonstrates the influence of different components of our
approach on the DAVIS validation set. First, we study the role of the
appearance stream. As a baseline, we remove it completely (``no'' in the ``App
stream'' in the table), i.e., the output of the motion stream is the only input
to our visual memory module. In this setting, the memory module lacks
sufficient information to produce accurate segmentations, which results in an
26.6\% drop in performance compared to the method where the appearance stream
with fc6 features is used (``Ours'' in the table). We then provide raw RGB
frames, concatenated with the motion prediction, as input to the ConvGRU. This
simplest form of image representation leads to a 14.8\% improvement, compared
to the motion only model, showing the importance of the appearance features.
The variant where RGB input is passed through two convolutional layers,
interleaved with $\tanh$ nonlinearities, that are trained jointly with the
memory module (``2-layer CNN''), further improves this. This shows the
potential of learning appearance representation as a part of the video
segmentation pipeline. Finally, we compare features extracted from the fc7 and
conv5 layers of the DeepLab model to those from fc6 used by default in our
method. Features from fc7 and fc6 show comparable performance, but fc7 ones are
more expensive to compute. Conv5 features perform significantly worse, perhaps
due to a smaller field of view.

The importance of appearance network pretrained on the semantic segmentation
task is highlighted by the ``ImageNet only'' variant in Table~\ref{tbl:abl},
where the PASCAL VOC pretrained segmentation network is replaced with a network
trained on ImageNet classification. Although ImageNet pretraining provides a
rich feature representation, it is less suitable for the video object
segmentation task, which is confirmed by an 6\% drop in performance. Discarding
the motion information (``no'' in ``Motion stream''), although being 10.5\%
below our complete method, still outperforms most of the motion-based
approaches on DAVIS (see Table~\ref{tbl:soadavis}). This variant learns
foreground/background segmentation, which is sufficient for videos with a
single dominant object, but fails in more challenging cases.

\begin{table*}[t]
\begin{center}
\begin{tabular}{c|c|c c c c c c c }
\hline
Measure & Set & KEY~\cite{lee2011key} & MP-Net-F~\cite{tokmakov2016learning} & FST~\cite{papazoglou2013fast} & CVOS~\cite{taylor2015causal} & CUT~\cite{keuper2015motion} & MP-Net-V~\cite{tokmakov2016learning} & Ours \\
\hline
\multirow{2}{*}{$\mathcal{P}$} & Training & 64.9 & 83.0 & 71.3 & 79.2 & 86.6 & 69.3 & 90.7 \\
& Test & 62.3 & 84.0 & 76.3 & 83.4 & 83.1 & 81.4 & 92.1 \\
\hline
\multirow{2}{*}{$\mathcal{R}$} & Training & 52.7 & 54.2 & 70.6 & 79.0 & 80.3 & 80.8 & 71.3 \\
 & Test & 56.0 & 49.4 & 63.3 & 67.9 & 71.5 & 73.9 & 67.4 \\
\hline
\multirow{2}{*}{$\mathcal{F}$} & Training & 58.2 & 65.6 & 71.0 & 79.3 & 83.4 & 74.6 & 79.8 \\
 & Test & 59.0 & 62.2 & 69.2 & 74.9 & 76.8 & 77.5 & 77.8 \\
\hline
\end{tabular}
\vspace{0.1cm}
\caption{Comparison to state-of-the-art methods on FBMS with precision ($\mathcal{P}$), recall ($\mathcal{R}$), and F-measure ($\mathcal{F}$).}
\label{tbl:bms}
\vspace{-0.5cm}
\end{center}
\end{table*}

Next, we evaluate the design choices in the visual memory module. Using a
simple recurrent model (ConvRNN) results in a slight decrease in performance.
Such simpler architectures can be used in case of a memory vs segmentation
quality trade off. The other variant using ConvLSTM is comparable to ConvRNN,
possibly due to the lack of sufficient training data. We also evaluated the
influence of the memory module (ConvGRU) by replacing it with a stack of 6
convolutional layers to obtain a memoryless variant of our model (``no'' in
``Memory module'' in Table~\ref{tbl:abl}), but with the same number of
parameters. This variant results in a 6\% drop in performance compared to our
full model. The performance of the memoryless variant is comparable
to~\cite{tokmakov2016learning} (63.3), the approach without any memory.
Performing a unidirectional processing instead of a bidirectional one decreases
the performance by nearly 3\% (``no'' in ``Bidir processing'').

Lastly, we train two variants (``FT3D GT Flow'' and ``FT3D LDOF Flow'') on the
synthetic FT3D dataset~\cite{Mayer16} instead of DAVIS. Both of them show a
significantly lower performance than our method trained on DAVIS. This is due
to the appearance of synthetic FT3D videos being very different from the
real-world ones. The variant trained on ground truth flow (GT Flow) is inferior
to that trained on LDOF flow because the motion network (MP-Net) achieves a
high performance on FT3D with ground truth flow, and thus our visual memory
module learns to simply follow the motion stream output.

\subsection{Comparison to MP-Net variants}
\label{sec:mpc}
In Table~\ref{tbl:mpn} we compare our method to MP-Net and its variants
presented in~\cite{tokmakov2016learning} on the DAVIS validation set.  Our
visual memory-based approach (``Ours'' in the table) outperforms the MP-Net
baseline (``MP-Net''), which serves as the motion stream in our model, by
16.5\%. This clearly demonstrates the value of the appearance stream and our
memory unit for video segmentation. The post-processing variants
in~\cite{tokmakov2016learning}, using objectness cues, CRF, and video
segmentation method~\cite{papazoglou2013fast}, improve this baseline, but
remain inferior to our result. Our full method (``Ours + CRF'') is nearly 6\%
better than ``MP-Net-Frame'', which is the best performing MP-Net variant
on DAVIS. 
Note that ``MP-Net-Video'' which combines MP-Net with
objectness cues and the video segmentation method of~\cite{papazoglou2013fast}
is also inferior to our method, as it relies strongly on the tracking
capabilities of~\cite{papazoglou2013fast}, which is prone to segmentation {\it
leaking} in case of errors in the flow estimation. The example in the first row in
Figure~\ref{fig:davis} shows a typical error of~\cite{papazoglou2013fast}.

MP-Net-Video performs better than MP-Net-Frame on the FBMS dataset (see
Table~\ref{tbl:bms}) since the frame-only variant does not segment objects when
they stop moving. The propagation of segment(s) over time with tracking in
MP-Net-Video addresses this, but is less precise due to segmentation leaks, as
shown by the comparison with precision measure in the table and the qualitative
results in Figure~\ref{fig:fbms}.

\subsection{Comparison to the state-of-the-art}
\label{sec:soa}
\vspace{0.3cm}\noindent\textbf{DAVIS.}
Table~\ref{tbl:soadavis} compares our approach to the state-of-the-art methods
on DAVIS. In addition to comparing our results to the top-performing
unsupervised approaches reported in~\cite{Perazzi16}, we evaluated two more
recent methods: CUT~\cite{keuper2015motion} and PCM~\cite{Bideau16}, with the
authors' implementation. Our method outperforms
MP-Net-Frame, the previous state of the art, by 5.9\% on the IoU measure, and
is 20.1\% better than the next best method~\cite{papazoglou2013fast}. We also
observe a 30.8\% improvement in temporal stability over MP-Net-Frame.
PCM~\cite{Bideau16}, which performs well on a subset of the FBMS dataset (as
shown in~\cite{tokmakov2016learning}), is in fact significantly worse on DAVIS.

\begin{figure}[t]
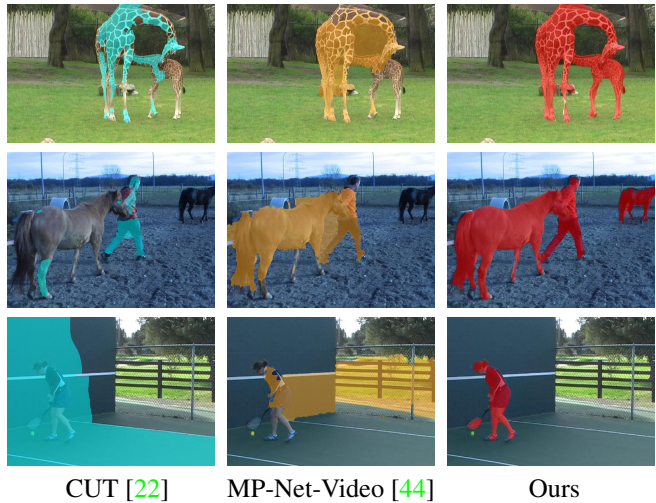

\begin{center}
\threefigures{soafbms/giraffes01/cut}{soafbms/giraffes01/mpnet}{soafbms/giraffes01/ours}{0.33\columnwidth}\vspace{0.1cm}
\threefigures{soafbms/horses04/cut}{soafbms/horses04/mpnet}{soafbms/horses04/ours}{0.33\columnwidth}\vspace{0.1cm}
\threefigurescaption{soafbms/tennis/cut}{soafbms/tennis/mpnet}{soafbms/tennis/ours}{0.33\columnwidth}{CUT~\cite{keuper2015motion}~~}{MP-Net-Video~\cite{tokmakov2016learning}}{~~~~~Ours}
\end{center}
\vspace{-0.3cm}\caption{Qualitative comparison with top-performing methods on
FBMS. Left to right: results of CUT~\cite{keuper2015motion},
MP-Net-Video~\cite{papazoglou2013fast}, and our method.}
\vspace{-0.4cm}
\label{fig:fbms}
\end{figure}
Figure~\ref{fig:davis} shows qualitative results of our approach, and the next
three top-performing methods on DAVIS:
MP-Net-Frame~\cite{tokmakov2016learning}, FST~\cite{papazoglou2013fast} and
CUT~\cite{keuper2015motion}. In the first row, both CUT and our method segment
the dancer, but our result is more accurate. FST leaks to the people in the
background and MP-Net misses parts of the person due to the incorrectly
estimated objectness. Our approach does not include any heuristics, which
makes it robust to this type of errors. In the second row, all the methods are
able to segment the swan, but only our method segments it completely and also
does not leak into the background. In the next row, our approach shows very
high precision, being able to correctly separate the dog and the pole occluding
it. In the last row, we illustrate a failure case of our method. The people in
the background move in some of the frames in this example. CUT, MP-Net and our
method segment them to varying extents. FST focuses on the foreground object,
but leaks to the background partially nevertheless.

\begin{figure*}[t]
\begin{center}
\begin{tabular}{lcc}
 & (a) {\it goat}, $t$ = 23 & (b) {\it dance-twirl}, $t$ = 19 \\
 & \includegraphics[width=0.4\linewidth]{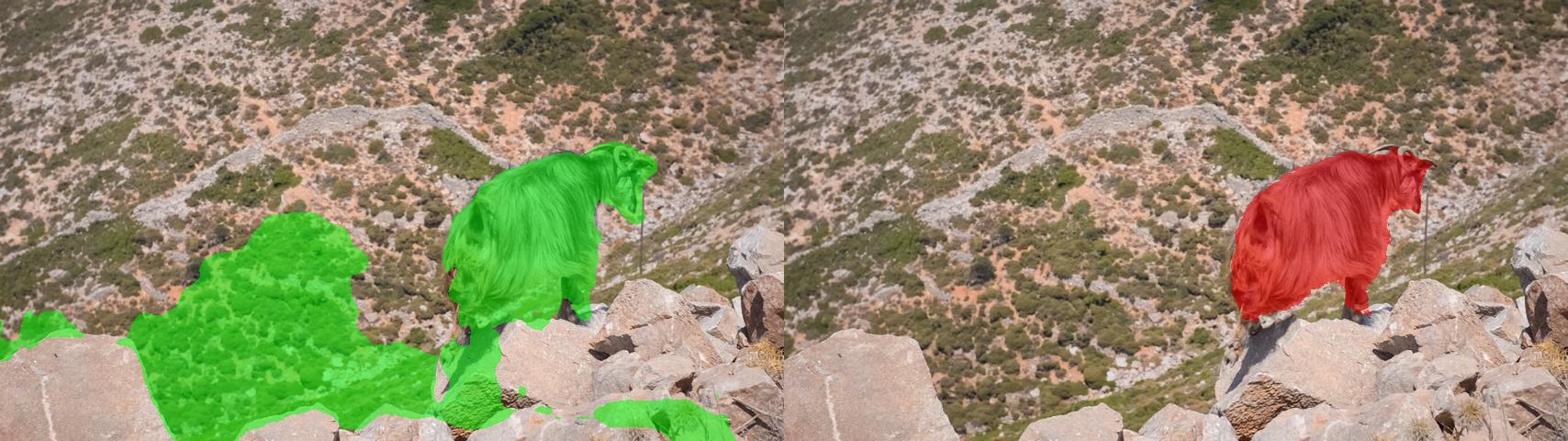} & \includegraphics[width=0.4\linewidth]{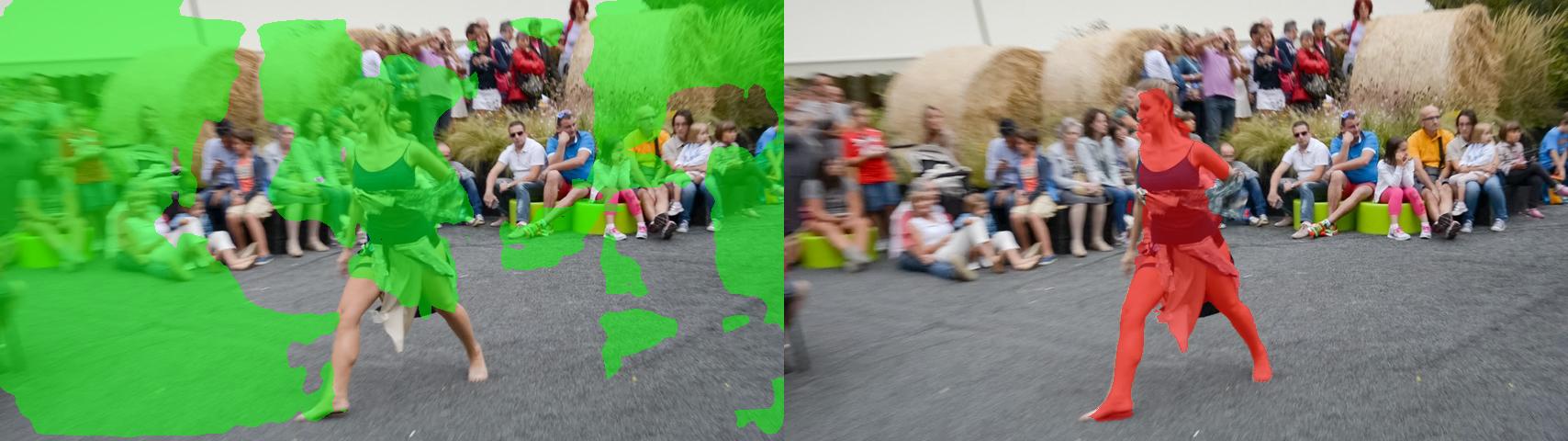}\\
$i =  8$ & \raisebox{-.4\height}{\includegraphics[width=0.4\linewidth]{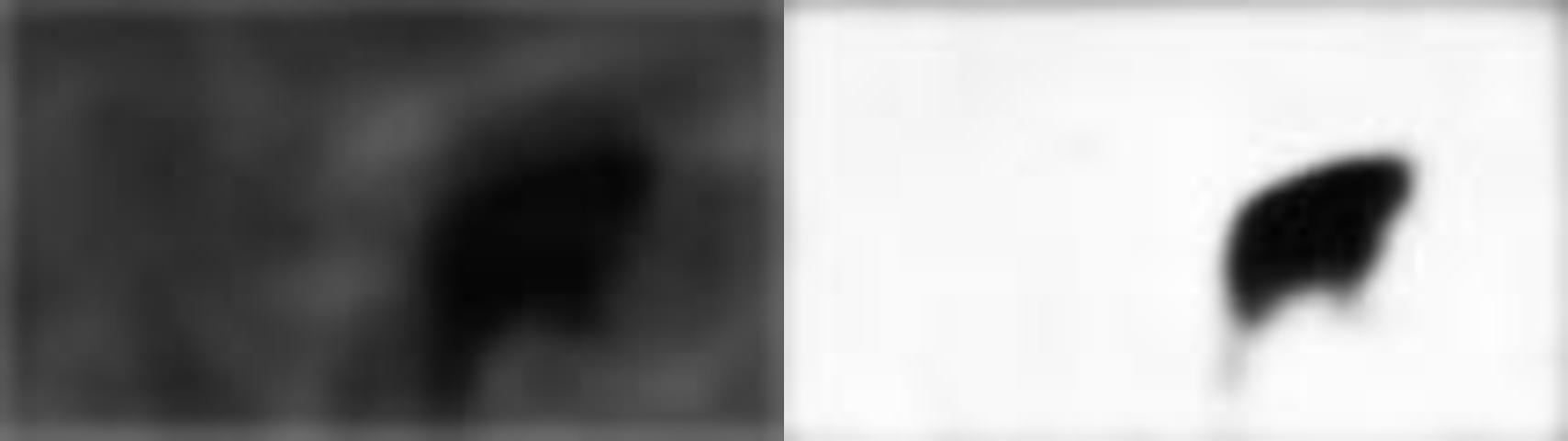}} & \raisebox{-.4\height}{\includegraphics[width=0.4\linewidth]{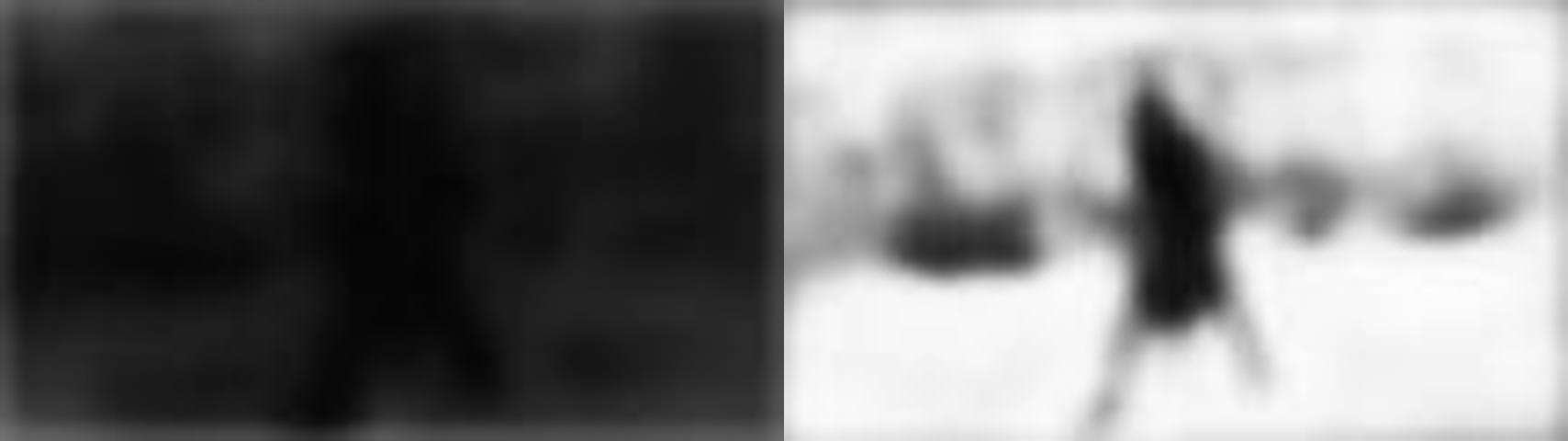}}\vspace{0.1cm}\\
$i = 18$ & \raisebox{-.4\height}{\includegraphics[width=0.4\linewidth]{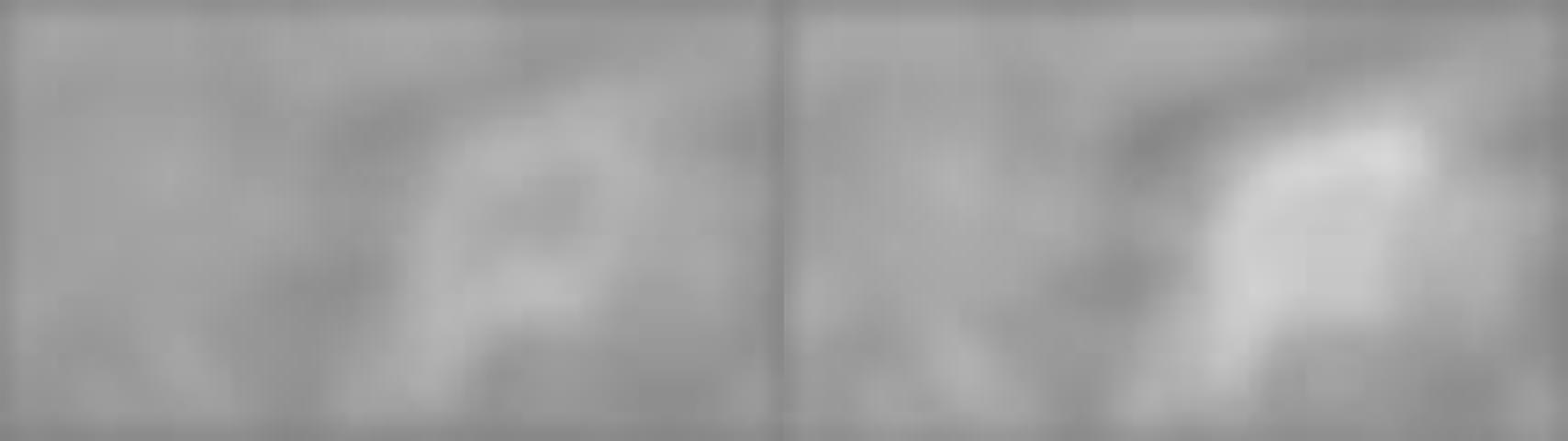}} & \raisebox{-.4\height}{\includegraphics[width=0.4\linewidth]{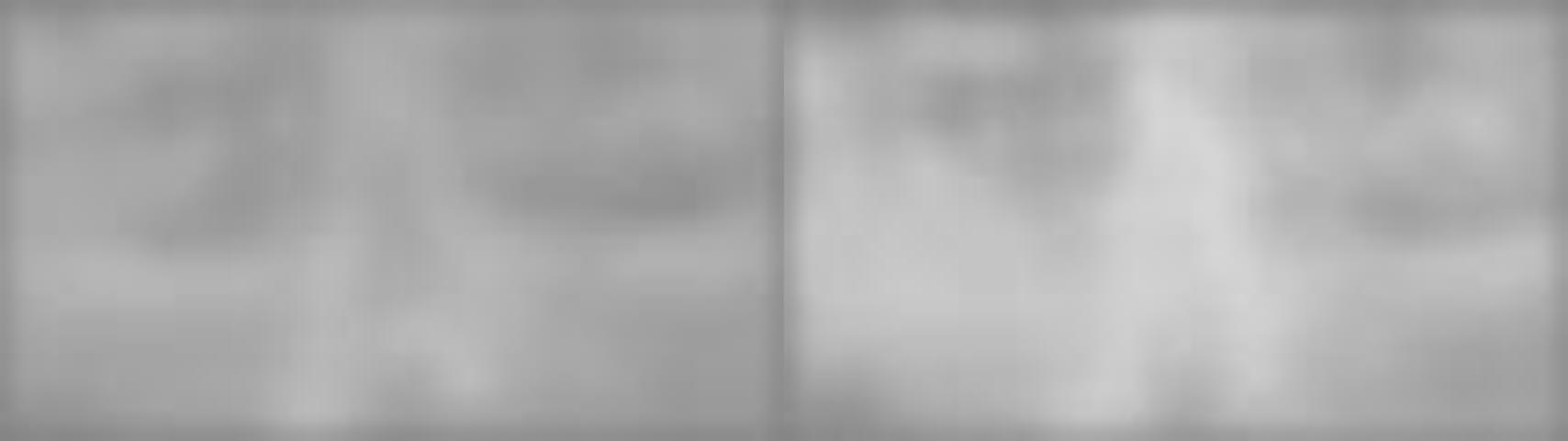}}\vspace{0.1cm}\\
$i = 28$ & \raisebox{-.4\height}{\includegraphics[width=0.4\linewidth]{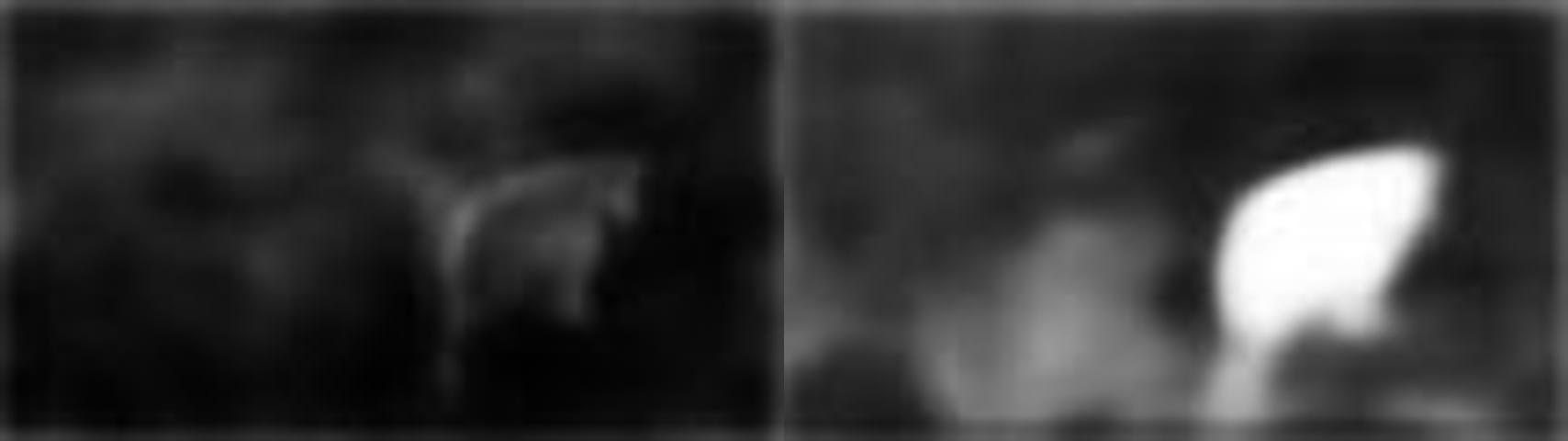}} & \raisebox{-.4\height}{\includegraphics[width=0.4\linewidth]{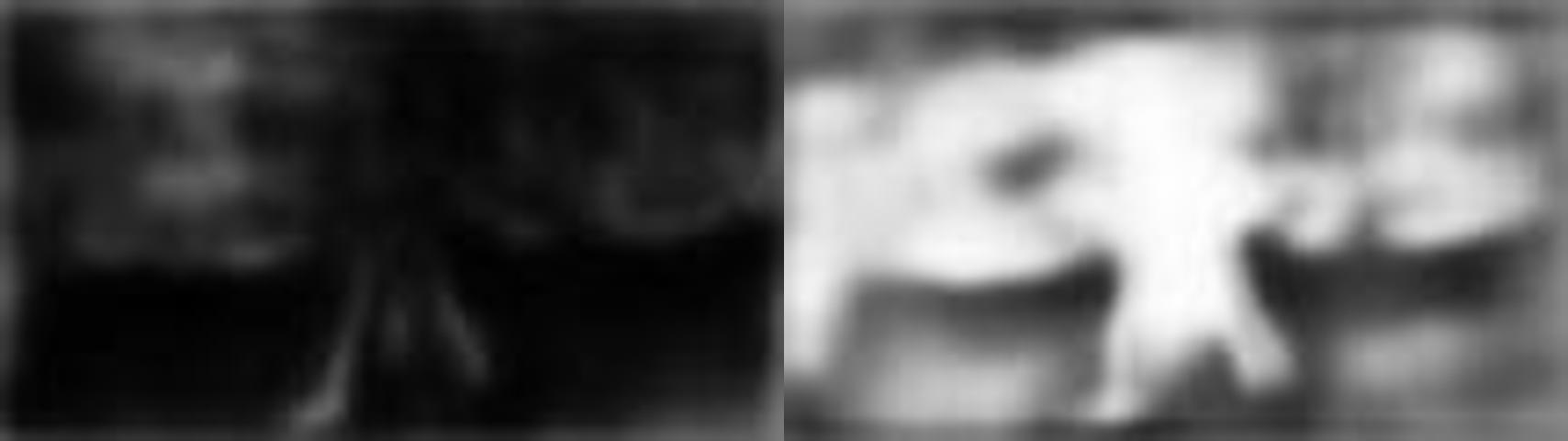}}\vspace{0.1cm}\\
$i = 41$ & \raisebox{-.4\height}{\includegraphics[width=0.4\linewidth]{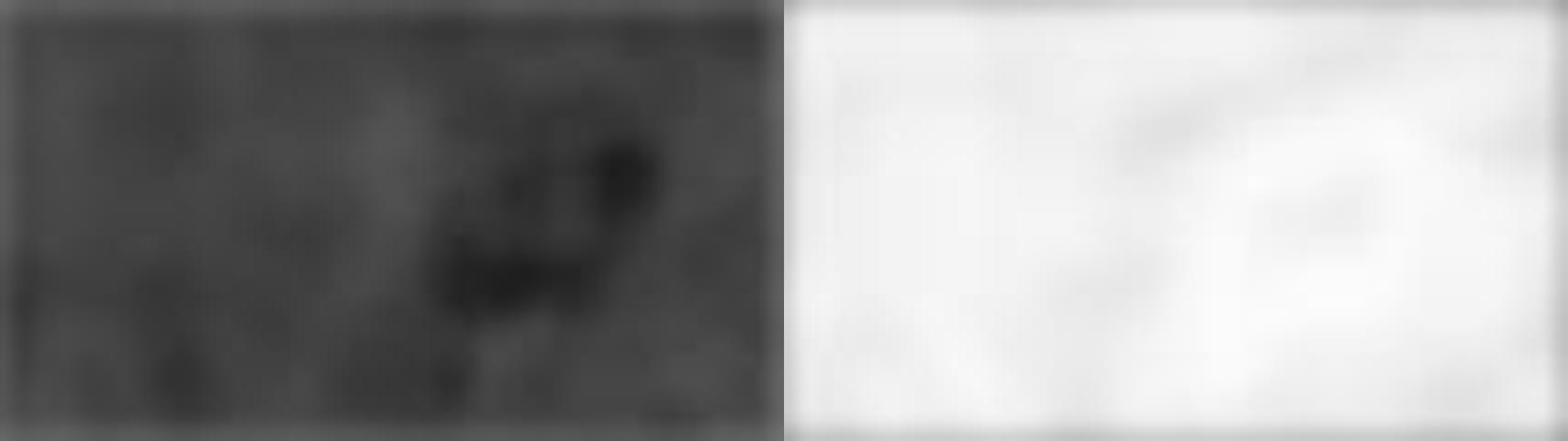}} & \raisebox{-.4\height}{\includegraphics[width=0.4\linewidth]{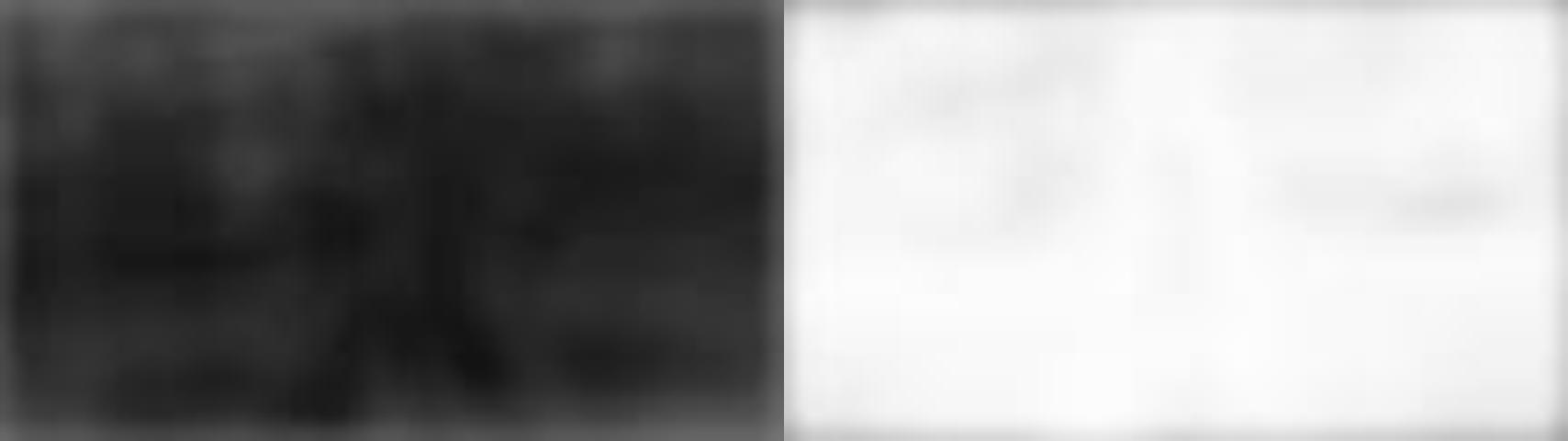}}\vspace{0.1cm}\\
$i = 63$ & \raisebox{-.4\height}{\includegraphics[width=0.4\linewidth]{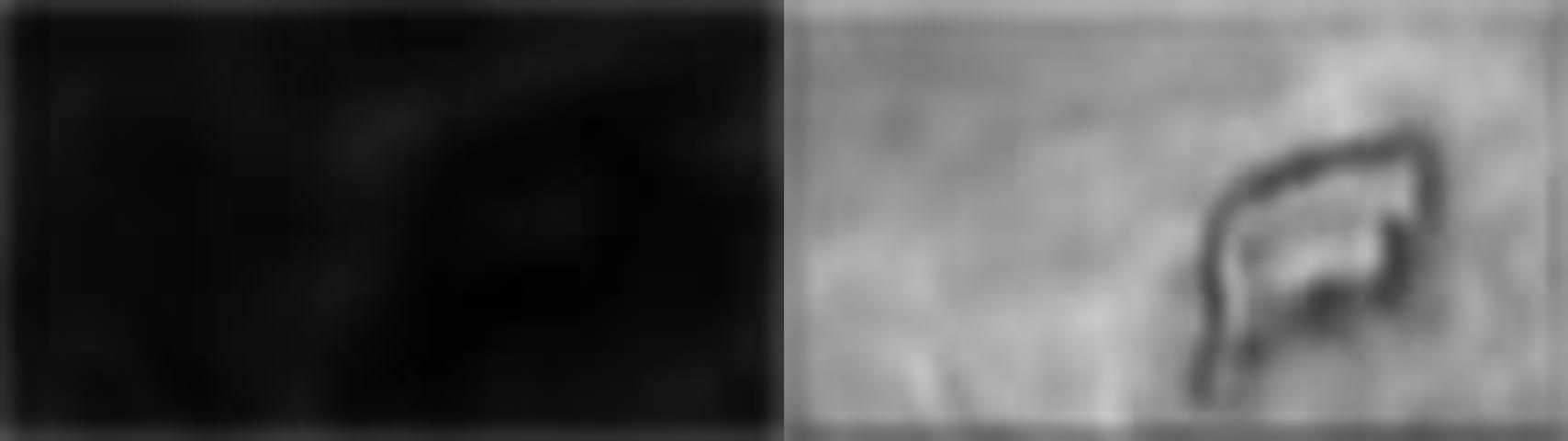}} & \raisebox{-.4\height}{\includegraphics[width=0.4\linewidth]{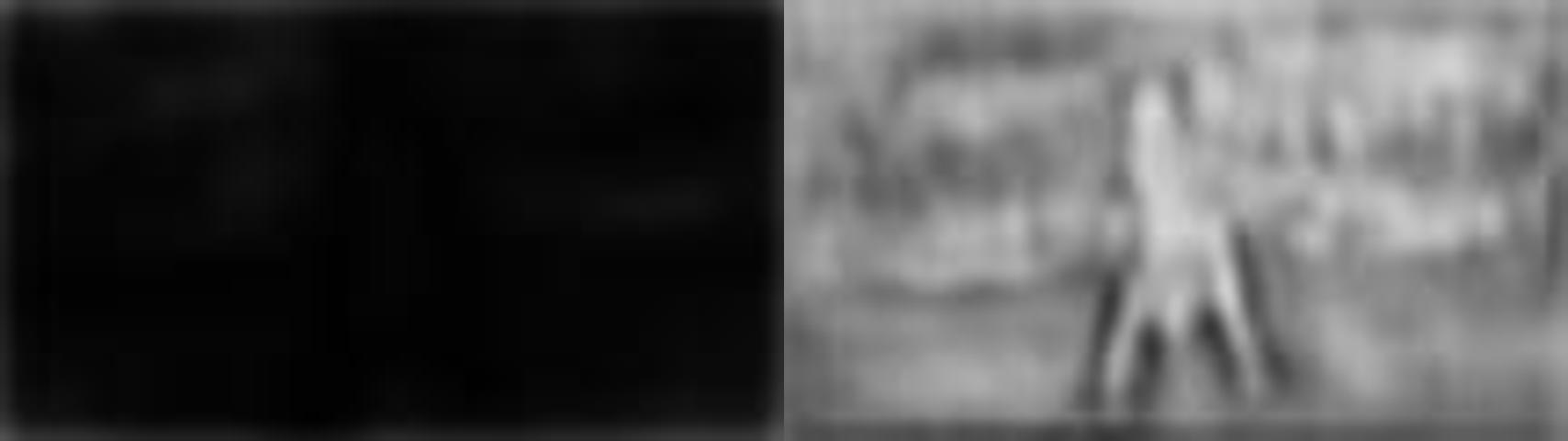}}\vspace{0.1cm}\\
\end{tabular}
\makebox[1\linewidth]{~~~~~~~~~~~~~~~~~~~~~~$r_t^i$ ~~~~~~~~~~~~~~~~~~~~~~~~~~~~~~~~$1 - z_t^i$  ~~~~~~~~~~~~~~~~~~~~~~~~~~~~~~~~~~~~$r_t^i$ ~~~~~~~~~~~~~~~~~~~~~~~~~~~~~~~~~$1 - z_t^i$ }
\end{center}
\caption{Visualization of the ConvGRU gate activations for two sequences from
the DAVIS validation set. The first row in each example shows the motion stream output
and the final segmentation result. The other rows are the reset
($r_t$) and the inverse of the update $(1 - z_t)$ gate activations for the corresponding $i$th
dimension. These activations are shown as grayscale heat maps, where white
denotes a high activation.}
\vspace{-0.4cm}
\label{fig:gru}
\end{figure*}

\vspace{0.3cm}\noindent\textbf{FBMS.}
As shown in Table~\ref{tbl:bms} MP-Net-Frame~\cite{tokmakov2016learning} is
outperformed by most of the methods on this dataset. Our approach based on
visual memory outperforms MP-Net-Frame by 15.6\% on the test set and by 14.2\%
on the training set according to the F-measure. FST~\cite{papazoglou2013fast}
based post-processing (``MP-Net-V'' in the table) significantly improves the
results of MP-Net on FBMS, but it remains below our approach on both precision
and F-measure. Overall, our method shows top results in terms of precision and
F-measure but is outperformed by some methods on recall. This is due to very
long static sequences present in FBMS, which our recurrent memory-based method
can not handle as well as methods with explicit tracking components, such as
CUT~\cite{keuper2015motion}.

Figure~\ref{fig:fbms} shows qualitative results of our method and the two
next-best methods on FBMS: MP-Net-Video~\cite{tokmakov2016learning} and
CUT~\cite{keuper2015motion}. MP-Net-Video relies highly on
FST's~\cite{papazoglou2013fast} tracking capabilities, and thus demonstrates
the same background leaking failure mode, as seen in all the three examples.
CUT misses parts of objects and incorrectly assigns background regions to the
foreground in some cases, whereas our method demonstrates very high precision.

\vspace{0.3cm}\noindent\textbf{SegTrack-v2.}
Our approach achieves IoU of 57.3 on this dataset. The relatively lower IoU
compared to DAVIS is mainly due to the low resolution of some of the
SegTrack-v2 videos, which differ from the high resolution ones used for
training. We have also evaluated the state-of-the-art approaches for comparison
on SegTrack. As shown in Table~\ref{tbl:strck}, our method performs better
than~\cite{keuper2015motion,papazoglou2013fast} on SegTrack, but worse than
NLC~\cite{Faktor14}. Note that NLC was designed and evaluated on SegTrack; we
outperform it on DAVIS by 20.8\% (see Table~\ref{tbl:soadavis}).
\begin{table}[t]
\begin{center}
\begin{tabular}{c c c c c}
\hline
CUT~\cite{keuper2015motion} & FST~\cite{papazoglou2013fast} &  NLC~\cite{Faktor14}  & Ours \\
\hline
47.8 & 54.3  & 67.2 & 57.3  \\
\hline
\end{tabular}
\end{center}
\vspace{0.1cm}
\caption{Comparison to state-of-the-art methods on SegTrack-v2 with mean IoU.}
\label{tbl:strck}
\end{table}

\subsection{ConvGRU visualization}
\label{sec:gru}
We present a visualization of the gate activity in our ConvGRU unit on two
videos from the DAVIS validation set. We use the unidirectional model in the
following for better clarity. The reset and update gates of the ConvGRU, $r_t$
and $z_t$ respectively, are 3D matrices of $64 \times h/8 \times w/8$
dimension.  The overall behavior of ConvGRU is determined by the interplay of
these 128 components. We use a selection of the components of $r_t$
and $(1 - z_t)$ to interpret the workings of the gates. Our analysis is shown on two frames which correspond
to the middle of the {\it goat} and {\it dance-twirl} sequences in (a) and (b)
of Figure~\ref{fig:gru}.

The outputs of the motion stream alone (left) and the final segmentation result
(right) of the two examples are shown in the top row in the figure. The five
rows below correspond each to one of the 64 dimensions of $r_t$ and $(1 -
z_t)$. These activations are shown as a grayscale heat map. High
values for either of the two activations increases the influence of the previous
state of a ConvGRU unit in the computation of the new state matrix. If
both values are low, the state in the corresponding locations is
rewritten with a new value; see equations (\ref{eqn:candmem}) and 
(\ref{eqn:state}).

For $i=8$, we observe the update gate being selective based on the appearance
information, i.e., it updates the state for foreground objects and duplicates
it for the background. Note that motion does not play a role in this case. This
can be seen in the example of stationary people (in the background) on the right, that are treated as foreground by the update gate. In the
second row, showing responses for $i=18$, both heatmaps are uniformly close to
$0.5$, which implies that the new features for this dimension are obtained by
combining the previous state and the input at the time step $t$.

In the third row for $i=28$, the update gate is driven by motion. It keeps the
state for regions that are predicted as moving, and rewrites it for other
regions in the frame. For the fourth row, where $i=41$, $r_t$ is uniformly
close to 0, whereas $(1 - z_t)$ is close to 1. As a result, the input is
effectively ignored and the previous state is duplicated. In the last row
showing $i=63$, a more complex behavior can be observed, where the gates
rewrite the memory for regions in object boundaries, and use both the previous
state and the current input for other regions in the frame.

\section{Conclusion}
This paper introduces a novel approach for video object segmentation. Our
method combines two complementary sources of information: appearance and
motion, with a visual memory module, realized as a bidirectional convolutional
gated recurrent unit. The ConvGRU module encodes spatio-temporal evolution of
objects in a video and uses this encoding to improve motion segmentation. The
effectiveness of our approach is validated on the DAVIS and FBMS datasets,
where it shows top performance. Instance-level video object segmentation is a
promising direction for future work.

\vspace{0.2cm}
\noindent {\bf Acknowledgments.}
This work was supported in part by the ERC advanced grant ALLEGRO, the
MSR-Inria joint project, a Google research award and a Facebook gift. We
gratefully acknowledge the support of NVIDIA with the donation of GPUs used for
this research.

{\small
\bibliographystyle{ieee}
\bibliography{egbib}
}

\end{document}